\documentclass{svproc}
\usepackage[utf8]{inputenc}
\usepackage{graphicx}
\usepackage{comment}
\usepackage{amsmath,amssymb} 
\usepackage{color}
\usepackage{xcolor}
\usepackage{hyperref}
\usepackage{float}
\usepackage{subfig}
\usepackage{subfiles} 
\usepackage{hyperref}

\title{Large-scale Neural Solvers for Partial Differential Equations}
\author{Patrick Stiller $^{1,2}$ \and Friedrich Bethke $^{1,2}$\and  Maximilian Böhme$^{3}$ \and Richard Pausch$^{1}$ \and Sunna Torge $^{2}$ \and Alexander Debus$^{1}$ \and Jan Vorberger$^{1}$ \and Michael Bussmann $^{3,1}$ \and Nico Hoffmann $^{1}$}

\institute{Helmholtz-Zentrum Dresden-Rossendorf, Dresden, Germany \and Technische Universität Dresden, Dresden, Germany \and Center for Advanced Systems Understanding (CASUS), Görlitz, Germany}
\date{08 June 2020}

\begin{document}

\maketitle

\section*{Abstract}
Solving partial differential equations (PDE) is an indispensable part of many branches of science as many processes can be modelled in terms of PDEs. However, recent numerical solvers require manual discretization of the underlying equation as well as sophisticated, tailored code for distributed computing. Scanning the parameters of the underlying model significantly increases the runtime as the simulations have to be cold-started for each parameter configuration. Machine Learning based surrogate models denote promising ways for learning complex relationship among input, parameter and solution. However, recent generative neural networks require lots of training data, i.e. full simulation runs making them costly. In contrast, we examine the applicability of continuous, mesh-free neural solvers for partial differential equations, physics-informed neural networks (PINNs) solely requiring initial/boundary values and validation points for training but no simulation data. The induced curse of dimensionality is approached by learning a domain decomposition that steers the number of neurons per unit volume and significantly improves runtime. Distributed training on large-scale cluster systems also promises great utilization of large quantities of GPUs which we assess by a comprehensive evaluation study. Finally, we discuss the accuracy of GatedPINN with respect to analytical solutions- as well as state-of-the-art numerical solvers, such as spectral solvers.
\section{Introduction}
Scientific neural networks accelerate scientific computing by data-driven methods such as physics-informed neural networks. One such prominent application is surrogate modelling which is e.g. used in particle physics at CERN\cite{Vallecorsa2018}. Enhancing neural networks by prior knowledge about the system makes the prediction more robust by regularizing either the predictions or the training of neural networks. One such prominent approach is a physics-informed neural network (PINN) which makes use of either learning\cite{Raissi2018} or encoding the governing equations of a physical system into the loss function\cite{Raissi2017} of the training procedure. Surrogate models based on PINN can be seen as a \textit{neural solvers} as the trained PINN predicts the time-dependent solution of that system at any point in space and time. Encoding the governing equations into the training relies on automatic differentiation (AD) as it is an easy computing scheme for accessing all partial derivatives of the system. However, AD also constrains the neural network architecture to use $C^{k+1}$ differentiable activation functions provided the highest order of derivatives in the governing system is $k$. Furthermore, the computational cost increases with the size of the neural network as the whole computational graph has to be evaluated for computing a certain partial derivative.
The main contribution of this paper is three-fold. First, we introduce a novel 2D benchmark dataset for surrogate models allowing precise performance assessment due to analytical solutions and derivatives. Second, we improve the training time by incorporating and learning domain decompositions into PINN. Finally, we conduct a comprehensive analysis of accuracy, power draw and scalability on the well known example of the 2D quantum harmonic oscillator. 

\section{Related Works}
Accelerated simulations by surrogate modelling techniques are carried out in two main directions. Supervised learning methods require full simulation data in order to train some neural network architecture, e.g. generative adversarial networks\cite{Vallecorsa2018} or autoencoders\cite{kim19deep}, to reproduce numerical simulations and might benefit from interpolation between similar configurations. The latter basically introduces a speedup with respect to numerical simulations, however generalization errors might challenge this approach in general. In contrast, self-supervised methods either embed neural networks within numerical procedures for solving PDE\cite{Tompson2017}, or incorporate knowledge about the governing equations into the loss of neural networks, so called physics-informed neural networks (PINN)\cite{Raissi2017}. The latter is can be seen as variational method for solving PDE. Finally, \cite{Raissi2018} demonstrated joint discovery of a system (supervised learning) and adapting to unknown regimes (semi-supervised learning). Recently, \cite{Shin2020} proved convergence of PINN-based solvers for parabolic and hyperbolic PDEs. Parareal physics-informed neural networks approach domain decomposition by splitting the computational domain into temporal slices and training a PINN for each slice\cite{PPINN}. We are going to generalize that idea by introducing conditional computing \cite{MoE} into the physics-informed neural networks framework, hereby enabling an arbitrary decomposition of the computational domain which is adaptively tuned during training of the PINN.
\newpage

\section{Methods}

The governing equations of a dynamic system can be modeled in terms of non-linear partial differential equations
\begin{equation*}
    u_t + \mathcal{N}(u; \lambda) = 0 \ ,
\end{equation*}
with $u_t = \frac{\partial u}{\partial t} $ being the temporal derivative of the solution $u$ of our system while $\mathcal{N}$ denotes a non-linear operator that incorporates the (non-)linear effects of our system. 
One example of such a system is the quantum harmonic oscillator, 
\begin{equation*}
    \label{eq:TDSE}
    i \frac{\partial \psi(\mathbf{r},t)}{\partial t} - \hat{H} \psi(\mathbf{r},t) = 0 \ ,
\end{equation*}
where $\psi(\mathbf{r},t)$ denotes the so-called state of the system in the spatial base and $\hat{H}$ is the Hamilton-operator of the system. The systems state absolute square $|\psi(\mathbf{r},t)|^2$ is interpreted as the probability density of measuring a particle at a certain point $\mathbf{r}$ in a volume $\mathcal{V}$. Thus, $|\psi(\mathbf{r},t)|^2$ has to fulfill the normalization constraint of a probability density
\begin{equation*}
\label{eqn:boundary}
    \int_{\mathcal{V }} d^3r \, |\psi(\mathbf{r},t)|^2 = 1. 
\end{equation*}

\begin{figure}[H]
    \centering
    \includegraphics[width=\textwidth]{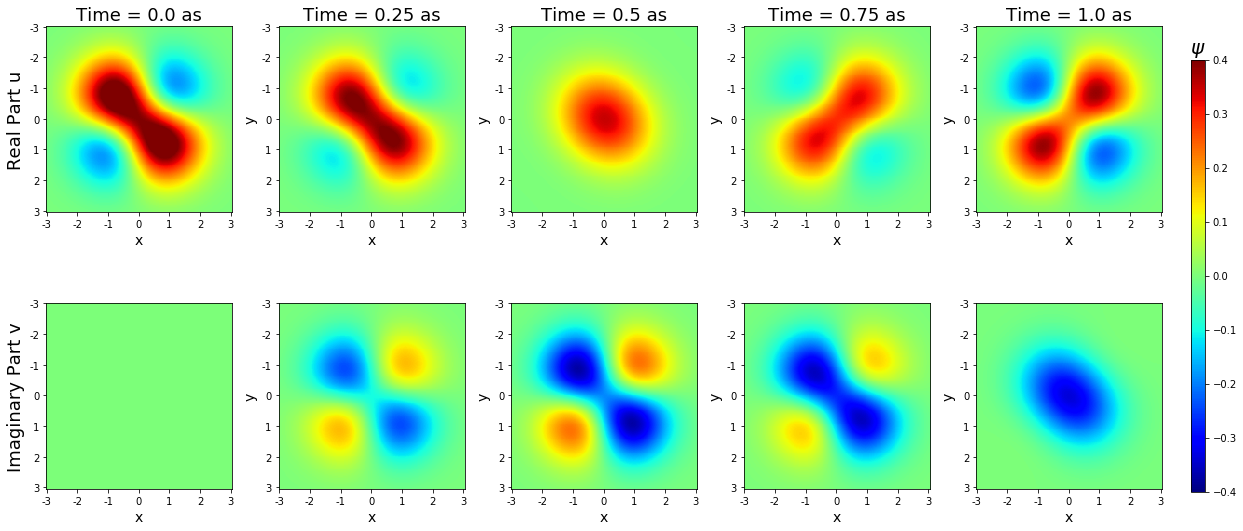}
    \caption{Analytic solution of the quantum harmonic oscillator}
    \label{fig:qho_exact}
\end{figure}

The Hamilton operator of a particle in an external potential is of the form
\begin{align*}
    \hat{H} = -\frac{1}{2}\Delta + V(\mathbf{r},t),
\end{align*}
where $\Delta$ is the Laplace operator and $V(\mathbf{r},t)$ is a scalar potential. The first term is the kinetic energy operator of the system and $V(\mathbf{r},t)$ its potential energy.
In this work, we use the atomic unit system meaning that $\hbar=m_e=1$. $\hat{H}$ is a Hermitian operator acting on a Hilbertspace $\mathcal{H}$. In this work we are focusing on the 2D quantum harmonic oscillator (QHO), which is described by the Hamiltonian
\begin{equation*}
\label{eqn:qho}
\hat{H} = -\frac{1}{2} \left( \frac{\partial^2}{\partial x^2} + \frac{\partial^2}{\partial y^2}  \right) + \frac{\omega_0^2}{2} (x^2+y^2) \\= \Hat{H}_x + \Hat{H}_y.
\end{equation*} 
where $x\in\mathbb{R}$ and $y\in\mathbb{R}$ denote spatial coordinates. The solution of the QHO can be determined analytically and is the basis for complicated systems like the density function theory (DFT). Therefore the QHO is very well suited as a test system which allows a precise evaluation of the predicted results. In addition, the QHO can also be used as a test system for evaluating the results. Furthermore, the QHO is classified as linear parabolic PDE, which guarantees the functionality of the chosen PINN approach according to Shin et al. \cite{Shin2020}. Figure \ref{fig:qho_exact} shows the analytic solution of the quantum harmonic oscillator over time. 

\subsection{Physics-informed Quantum Harmonic Oscillator}
The solution $\psi(x,y,t)$ of our quantum harmonic oscillator at some position $x,y$ and time $t$ is approximated by a neural network $f: \mathbb{R}^3 \to \mathbb{C}$, i.e. 
\begin{equation*}
\widehat{\psi}(x,y,t) = f(x,y,t) \ .
\end{equation*}
In this work, we model $f$ by a simple multilayer perceptron (MLP) of $1 \leq l \leq m$ layers, a predetermined number of neurons per layer $k_l$ and respective weight matrices $W^l\in\mathbb{R}^{k_l \times k_l}$
\begin{equation*}
y^l = g(W^ly^{l-1}) \ ,
\end{equation*}
with $y^0 = (x,y,t)$ and $y^m = \widehat{\psi}(x,y,t)$. The training of Physics-informed neural networks relies on automatic differentiation which imposes some constraints on the architecture. In our case, the network has to be 3 times differentiable due to the second-order partial derivatives in our QHO (eqn. \ref{eqn:qho}). This is achieved by choosing at least one activation function $g$ which fulfills that property (e.g. tanh).  The training of the neural network is realized by minimizing the combined loss $\mathcal{L}$ defined in equation \eqref{eq:pde-loss-qho}. The three terms of $\mathcal{L}$ relate to the error of representing the initial condition $L_{0}$, the fulfillment of the partial differential equation $L_{f}$ as well as boundary condition $L_{b}$.
\begin{equation}\label{eq:pde-loss-qho}
    \mathcal{L} =  \alpha L_{0}(\mathcal{T}_{0}) +  L_{f}(\mathcal{T}_{f}) + L_{b}(\mathcal{T}_{b})
\end{equation}

$\mathcal{L}_{0}$ is the summed error of predicted real- $u=real(\psi)$ and imaginary- $v=imag(\psi)$ of the initial state with respect to groundtruth real- $u^i$ and imaginary part $v^i$ at points $\mathcal{T_0}$. We introduce a weighting term $\alpha$ into $\mathcal{L}$ allowing us to emphasize the contribution of the initial state. 

 \begin{equation*}\label{eqn:L0}
 L_{0}(\mathcal{T}_{0})= \frac{1}{|\mathcal{T}_{0}|} \sum_{i=1}^{|\mathcal{T}_{0}|}\left|u\left(t_{0}^{i}, x_{0}^{i},y_{0}^{i}\right)-u^{i}\right|^{2} + \frac{1}{|\mathcal{T}_{0}|} \sum_{i=1}^{|\mathcal{T}_{0}|}\left|v\left(t_{0}^{i}, x_{0}^{i},y_{0}^{i}\right)-v^{i}\right|^{2}
\end{equation*}

The boundary conditions (eqn. \ref{eqn:boundary}) are modelled in terms of $L_b$ at predetermined spatial positions $T_b$ at time $t$.

\begin{equation*}\label{eq:l_b}
L_{b}\left(T_{b}, t\right)=1-\left(\iint_{T_{b}}\left(u(t,x,y)^{2}+v(t,x,y)^{2}\right) d x d y\right)^{2}
\end{equation*}

$\mathcal{L}_{f}$ is divided into real- and imaginary part, such that $f_u$ represents the correctness of the real- and $f_v$ the correctness of imaginary part of the predicted solution. This loss term is computed on a set $\mathcal{T}_{f}$ of randomly distributed \textit{residual points} that enforce the validity of the PDE at residual points $\mathcal{T}_f$.

\begin{equation*}
 L_{f}(\mathcal{T}_{f})= \frac{1}{|\mathcal{T}_{f}|} \sum_{i=1}^{|\mathcal{T}_{f}|}\left|f_u\left(t_{f}^{i}, x_{f}^{i},y_{f}^{i} \right)\right|^{2} + \frac{1}{|\mathcal{T}_{f}|} \sum_{i=1}^{|\mathcal{T}_{f}|}\left|f_v\left(t_{f}^{i}, x_{f}^{i},y_{f}^{i} \right)\right|^{2}
\end{equation*}

\begin{equation*}
\label{eqn:Lf}
f_u = - u_t - \frac{1}{2}v_{xx} - \frac{1}{2}v_{yy}+ \frac{1}{2}x^2v + \frac{1}{2}y^2v 
 \end{equation*}
 \begin{equation*}
f_v = - v_t + \frac{1}{2}u_{xx} + \frac{1}{2}u_{yy}- \frac{1}{2}x^2u - \frac{1}{2}y^2u 
 \end{equation*}

\subsection{GatedPINN}
Numerical simulations typically require some sort of domain decomposition in order to share the load among the workers. physics-informed neural networks basically consist of a single multilayer perceptron network $f$ which approximates the solution of a PDE for any input $(x,y,t)$. However, this also implies that the capacity of the network per unit volume of our compute domain increases with the size of the compute domain. This also implies that the computational graph of the neural network increases respectively meaning that the time and storage requirements for computing partial derivatives via automatic differentiation increases, too. This limits the capacity of recent physics-informed neural network. 

We will be tackling these challenges by introducing conditional computing into the framework of physics-informed neural networks. Conditional Computing denotes an approach that activate only some units of a neural network depending on the network input \cite{conditionalComputing}. A more intelligent way to use the degree of freedom of neural networks allows to increase the network capacity (degree of freedom) without an immense blow up of the computational time \cite{MoE}. \cite{PPINN} introduced a manual decomposition of the compute domain and found that the capacity of the neural network per unit volume and thus the training costs are reduced. However, this approach requires another coarse-grained PDE solver to correct predictions. 
A decomposition of the compute domain can be learned by utilizing the mixture of expert approach \cite{MoE} based on a predetermined number of so-called experts (neural networks). A subset $k$ of all $N$ experts are active for any point in space and time while the activation is determined by gating network which introduces an adaptive domain decomposition. The combination of mixture of experts and physics-informed neural networks leads to a new architecture called \textit{GatedPINN}. 

\subsubsection{Architecture}
\begin{figure}[H]
    \centering
    \includegraphics[width=\columnwidth]{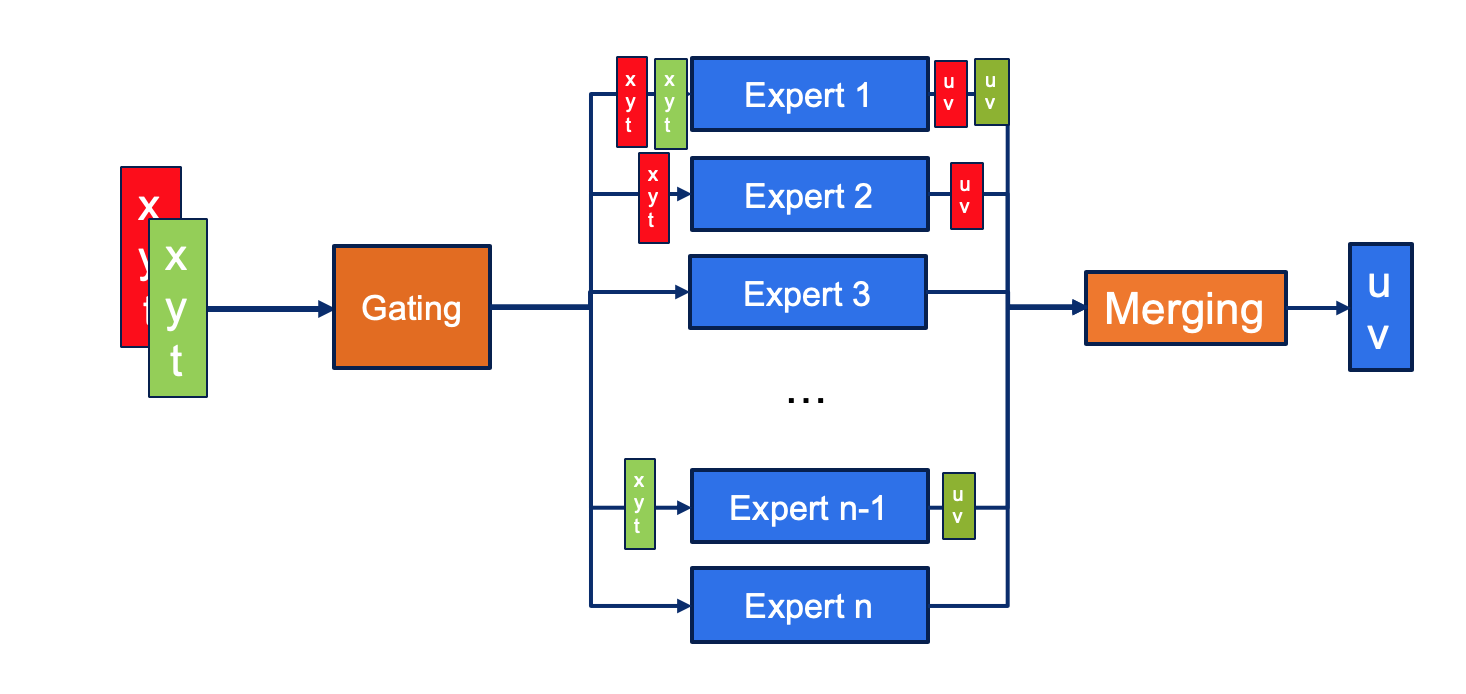}
    \caption{Visualization of the Gated-PINN architecture}
    \label{fig:gated_pinn_architecture}
\end{figure}

The architecture comprises of a gating network $G(x,y,t)$ that decides which expert $E_i(x,y,t)$ to use for any input $(x,y,t)$ in space and time (see Fig. \ref{fig:gated_pinn_architecture}). Experts $E_i$ with $1 \leq i \leq N$ are modelled by a simple MLP consisting of linear layers and tanh activation functions. The predicted solution $\widehat{\psi}$ of our quantum harmonic oscillator (QHO) becomes a weighted sum of expert predictions $E_i$

\begin{equation*}
    \widehat{\psi}(x,y,t) = \sum_{i=1}^{N}G(x,y,t)_i \cdot E_i(x,y,t) \ .
\end{equation*}

GatedPINN promise several advantages compared to the baseline PINN: First, the computation of partial derivatives by auto differentiation requires propagating information through a fraction $k/N$ of the total capacity of all experts. That allows to either increase the computational domain and/or increase the overall capacity of the neural network without a blow up in computational complexity. 
Similarly to \cite{MoE}, an importance loss $L_{I} = w_{\text{I}} \cdot CV(I(x,y,t))^2$ penalizes uneven distribution of workload among all $N$ experts:

\begin{equation}\label{eq:pde-loss-qho}
    L(\mathcal{T},\theta) =  L_{0}(\mathcal{T}_{0},\theta) +  L_{f}(\mathcal{T}_{f},\theta) + L_{b}(\mathcal{T}_{b},\theta) + \sum_{(x,y,t) \in T}L_{I}(X) \ ,
 \end{equation}
given $T = T_0 \cup T_b \cup T_f$. The importance loss $L_I(X)$ requires the computation of an importance measure $I(X) = \sum_{x \in X}G(x,y,t)$. The coefficient of variation $CV(z) = \sigma(z)/\mu(z)$ provided $I(X)$ quantifies the sparsity of the gates and thus the utilization of the experts. Finally, coefficient $w_{I}$ allows us to weight the contribution of our importance loss with respect to the PDE loss. The importance loss is defined as follows:

\begin{equation*}
    L_{I}(X)=w_{I} \cdot C V(\operatorname{I}(X))^{2} \ .
\end{equation*}

\subsubsection{Adaptive Domain Decomposition}
A trainable gating network $G$ allows us to combine the predictions of $k$ simple neural networks for approximating the solution of our QHO at any point in space $x,y$ and time $t$. Hereby, we restrict the size of the computational graph to $k$-times the size of each individual neural network $E^i$ with $0\leq i \leq k$.
\begin{equation*}
G(x,y,t) = \operatorname{Softmax}(\operatorname{KeepTopK}(H(x,y,t,\omega)))
\end{equation*}

and basically yields a $N$ dimensional weight vector with $k$ non-zero elements\cite{MoE}. The actual decomposition is learnt by the function $H$:
\begin{equation*}
\label{eqn:H_gating}
H(x,y,t) = ([x,y,t] \cdot W_g) + \operatorname{StandardNormal}() \cdot \operatorname{Softplus}(([x,y,t] ^T \cdot W_{noise})) \ .
\end{equation*}

The noise term improves load balancing and is deactivated when using the model. Obviously, this gating results in a decomposition into linear subspaces due to $W_g$. Non-linear domain decomposition can now be realized by replacing the weight matrix $W_g$ by a simple MLP $NN_{g}$, i.e. $([x,y,t] \cdot W_g)$ becomes $NN_{g}(x,y,t)$. This allows for more general and smooth decomposition of our compute domain.

\section{Results}

 All neural networks were trained on the Taurus HPC system of the Technical University of Dresden. Each node consists of two IBM Power9 CPUs and is equipped with six Nvidia Tesla V-100 GPUs. We parallelized the training of the neural networks using Horovod\cite{hvd} running on MPI communication backend. Training of the Physics-informed neural network, i.e. solving our QHO, was done on batches consisting of 8.500 points of the initial condition (i.e. $|T_0|$), 2.500 points for the boundary condition(i.e. $|T_b|$) and 2 million residual points (i.e. $|T_f|$).
 
\subsection{Approximation quality}
Training of physics-informed neural networks can be seen as solving partial differential equations in terms of a variational method. State-of-the-art solvers for our benchmarking case, the quantum harmonic oscillator, make us of domain knowledge about the equation by solving in Fourier domain or using Hermite polynomials. We will be comparing both, state-of-the-art spectral method \cite{SpectralSolver} as well as physics-informed neural networks, to the analytic solution of our QHO. This enables a fair comparison of both methods and allows us to quantify the approximation error.

For reasons of comparison, we use neural networks with similar capacity. The baseline model consists of 700 neurons at 8 hidden layer. The GatedPINN with linear and nonlinear gating consists of $N=10$ experts while the input is processed by one expert($k=1$). The experts of the GatedPINN are small MLP with 300 neurons at 5 hidden layers. Furthermore, the gating network for the nonlinear gating is also a MLP. It consists of a single hidden layer with 20 neurons and the ReLu activation function. 

The approximation error is quantified in terms of the infinity norm:
\begin{equation}
    err_{\infty} = ||\widehat{\psi} - \psi||_{\infty} \ ,
\end{equation}
which allow us to judge the maximum error while not being prone to sparseness in the solution. The relative norm is used for quantifying the satisfaction of the boundary conditions. The relative norm is defined with the approximated surface integral and the sampling points from dataset $T_b$ as follows

\begin{equation}
    err_{rel} = ||1 - \iint_{T_b} \psi \ dx dy|| \cdot 100\% \ .
\end{equation}

\begin{table}[H]
    \centering
    \begin{tabular}{|c|c|c|c|}
        \hline
      Approach & $err_{\infty}$ & Min & Max \\ \hline
      Spectral Solver & \textbf{0.01562} $\pm$ 0.0023 & 5.3455e-7 & 0.0223\\
      \hline
      PINN & 0.0159 $\pm$ 0.0060 & 0.0074 & 0.0265\\
      Linear GatedPINN & 0.0180 $\pm$ 0.0058 & 0.0094 & 0.0275\\
      Nonlinear GatedPINN & 0.0197 $\pm$ 0.0057 & 0.0098 & 0.0286\\
      \hline
    \end{tabular}
    \caption{Real part statistics of the infinity norm}
    \label{tab:infinity_real}
\end{table}

\begin{table}[H]
    \centering
    \begin{tabular}{|c|c|c|c|}
        \hline
      Approach & $err_{\infty}$ & Min & Max \\ \hline
      Spectral Solver & 0.01456 $\pm$ 0.0038  & 0.0000 & 0.0247 \\
      \hline
      PINN & \textbf{0.0144} $\pm$0.0064&0.0034&0.0269\\
      Linear GatedPINN & 0.0164 $\pm$ 0.0069&0.0043&0.0296\\
      Nonlinear GatedPINN & 0.0167 $\pm$ 0.0066&0.0046&0.0291\\
      \hline
    \end{tabular}
    \caption{Imaginary part statistics of the infinity norm}
    \label{tab:infinity_imag}
\end{table}

\begin{figure}[H]
    \centering
    \subfloat[Real Part]{{\includegraphics[width=0.45\textwidth]{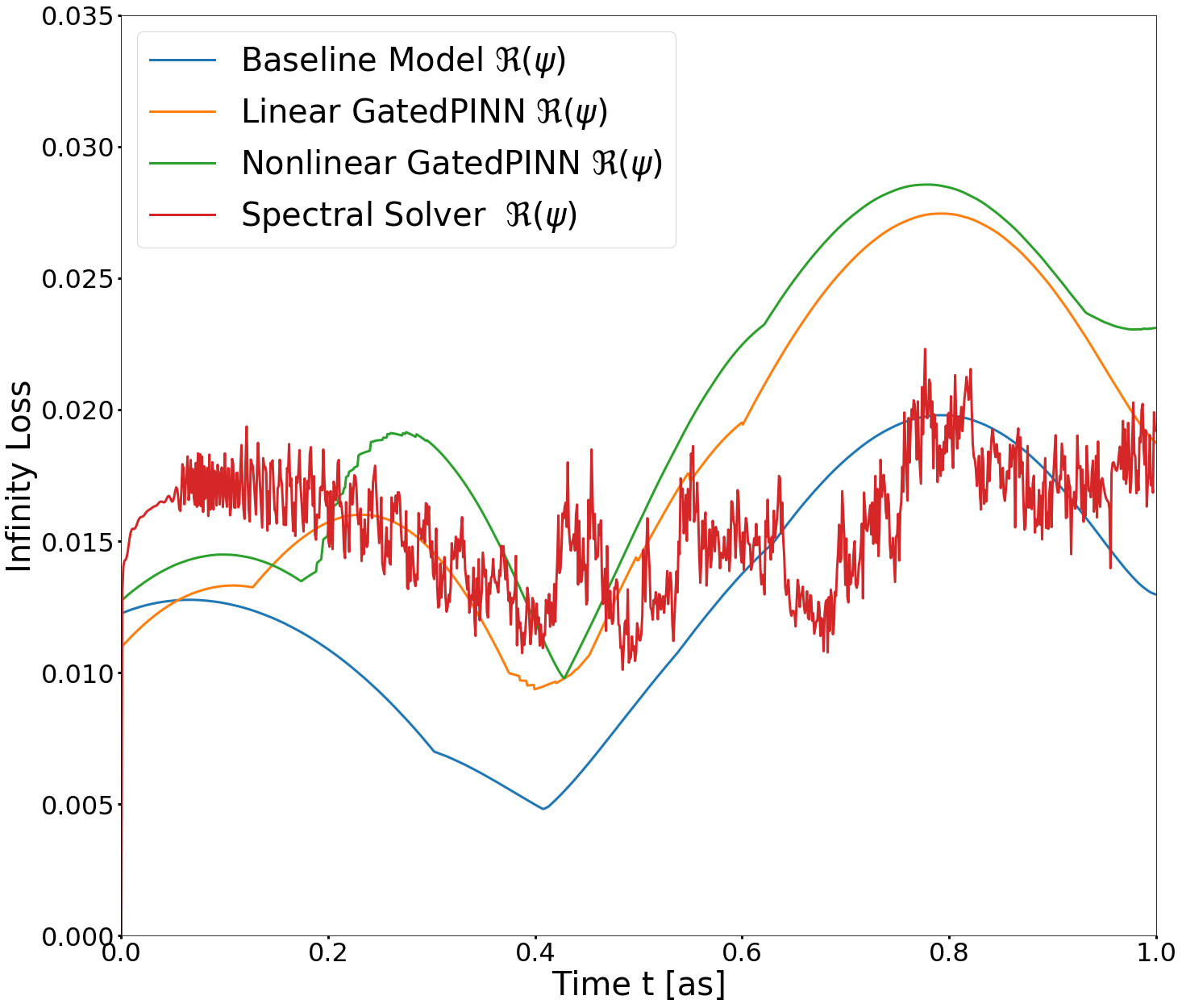}}}%
    \qquad
    \subfloat[Imaginary Part]{{\includegraphics[width=0.45\textwidth]{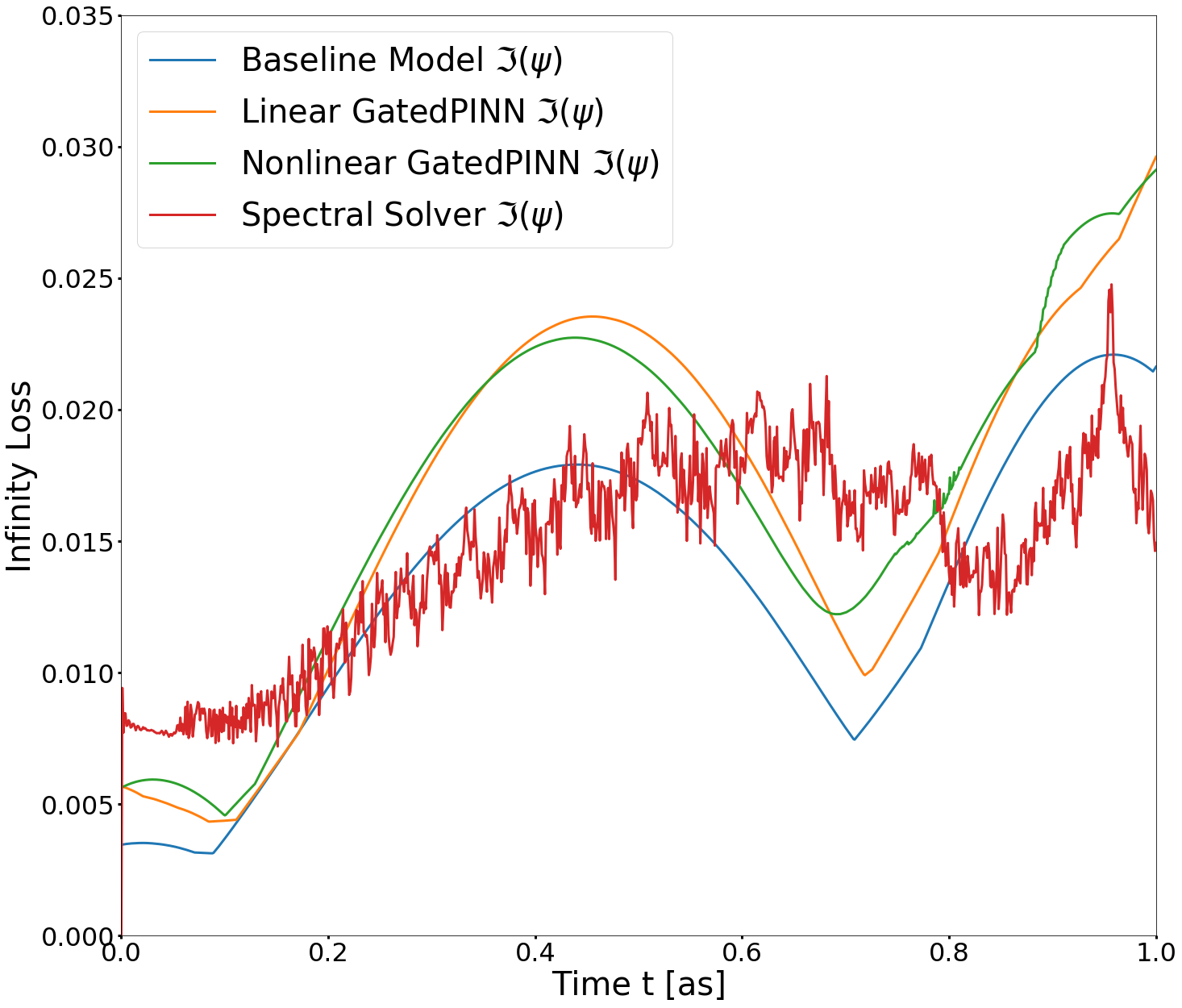}}}%
    \caption{Quality of the real part and imaginary part predictions over time in comparison to the spectral solver in reference to the analytically solution}
    \label{fig:infinity}
\end{figure}

Physics-informed neural networks as well as GatedPINN are competitive in quality to the spectral solver for the quantum harmonic oscillator in the chosen computational domain as can be seen in fig. \ref{fig:infinity}. The periodic development in the infinity norm relates to the rotation of the harmonic oscillator which manifests in the real as well as imaginary at different points in time (see Fig. \ref{fig:qho_exact}).

\begin{figure}[H]
    \centering
    \includegraphics[width=\textwidth]{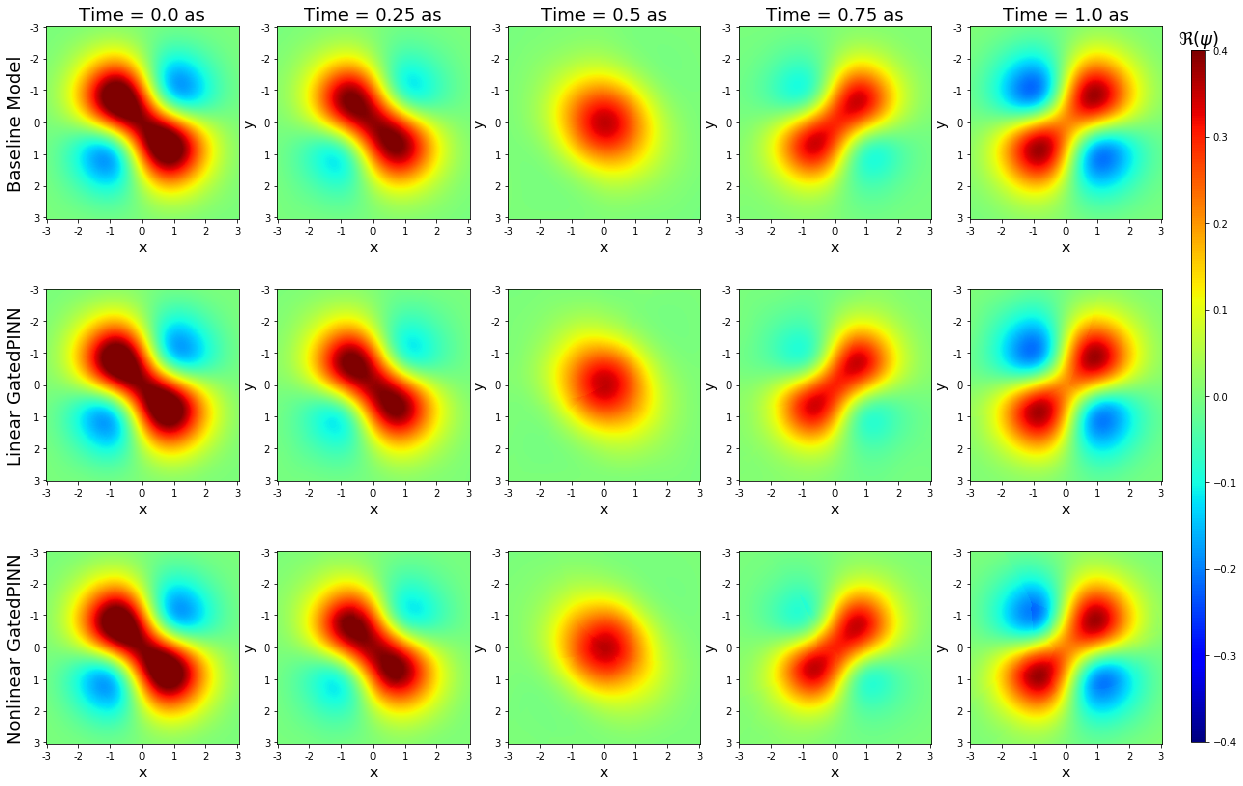}
    \caption{Real Part predictions of the Baseline and the GatedPINN models}
    \label{fig:inference_real_part}
\end{figure}

\begin{figure}[H]
    \centering
    \includegraphics[width=\textwidth]{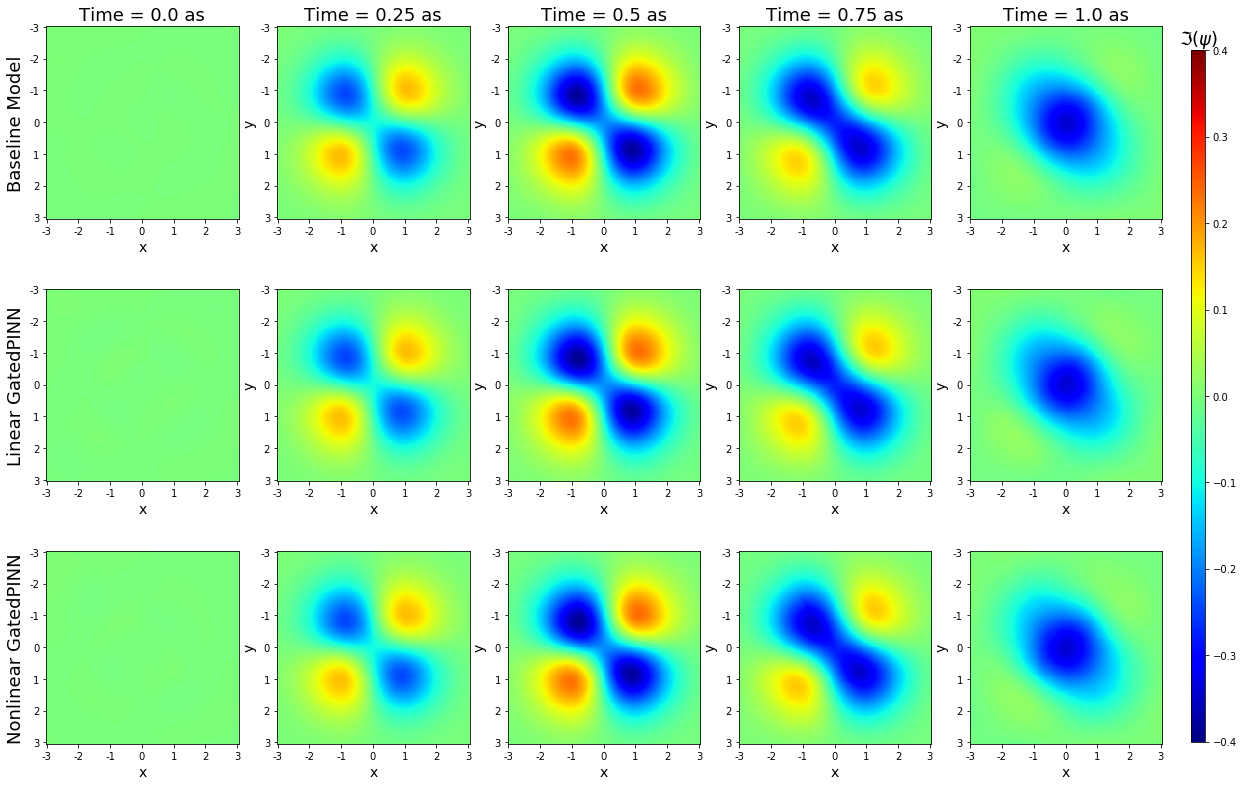}
    \caption{Imaginary Part predictions of the Baseline and the GatedPINN models}
    \label{fig:inference_imaginary_part}
\end{figure}

Fig. \ref{fig:inference_real_part} and Fig. \ref{fig:inference_imaginary_part} show the time evolution of the PINN predictions. The prediction of the baseline model and the GatedPINN models show the same temporal evolution as in Fig. \ref{fig:qho_exact}.

\subsection{Domain decomposition}
\begin{table}[H]
\centering
\begin{tabular}{|c|c|c|c|}
    \hline
      Model & Parameters &  $\mathcal{L}$ & Training Time \\ \hline
      PINN & \textbf{3,438,402} &  2.51e-4 & 29 h 19 min \\
      Linear GatedPINN & 3,627,050 & \textbf{2.115e-4} & \textbf{17 h 42 min} \\
      Nonlinear GatedPINN & 3,627,290 & 2.270e-4 & 18 h 08 min \\
      \hline
 \end{tabular}
 \caption{Training time of physics-informed neural networks is significantly reduced by incorporating a domain decomposition into the PINN framework.}
 \label{tab:Convergence_pde_loss}
\end{table}

Table \ref{tab:Convergence_pde_loss} shows the convergence of the PINN-Loss of the baseline, the GatedPINN with linear and nonlinear gating. The Baseline model and the GatedPINN models are trained with 2 million residual points and with the same training setup in terms of batch size, learning rate. Both, the GatedPINN with linear and nonlinear gating have converged to a slightly lower PINN-Loss as the baseline model. However, the training times of the Gated PINN are significantly shorter although the GatedPINN models have more parameters than the baseline model. These results show the efficient usage of the model capacity and automatic differentiation of the GatedPINN architecture. However, both the training time of the PINN and the GatedPINN approach is not competitive to the solution time of the spectral solver (1 min 15 sec). The full potential of PINN can only be used when they learn the complex relationship between the input, the simulation parameters and the solution of the underlying PDE and thus restarts of the simulation can be avoided. 

In table \ref{tab:infinity_real} and \ref{tab:infinity_imag} we see that the approximation quality of the baseline model is sligthly better than the GatedPINN models although the GatedPINN models have converged to a slightly smaller loss $\mathcal{L}$. However, the GatedPINN (linear: 0.329 \%, nonlinear: 0.268 \%) satisfies the boundary condition better than the baseline model (1.007 \%). This result could be tackled by introducing another weighting constant similarly to $\alpha$ to Eq. \ref{eq:pde-loss-qho}.

\begin{figure}[H]
    \centering
    \includegraphics[width=\textwidth]{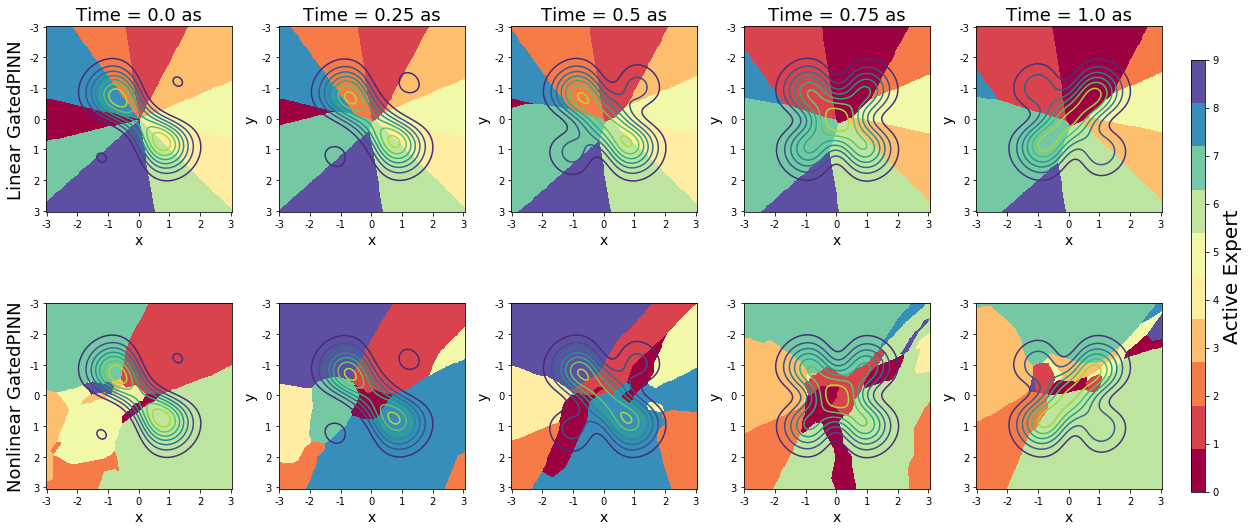}
    \caption{Learned domain decomposition by the GatedPINN with linear and non-linear gating. The squared norm of the solution $\psi$ is visualized as a contour plot}
    \label{fig:domain_decomposition}

\end{figure}

The learned domain decomposition of the proposed GatedPINN can be seen in Fig. \ref{fig:domain_decomposition}. The nonlinear gating, which is more computationally intensive, shows an more adaptive domain decomposition over time than the model with linear gating. The linear gating converges to a fair distribution over the experts. The nonlinear approach converges to a state where the experts are symmetrically distributed in the initial state. This distribution is not conserved in the time evolution. 

\subsection{Scalability \& power draw}
Training of neural solvers basically relies on unsupervised learning by validating the predicted solution $\psi$ on any residual point (Eq. \ref{eq:pde-loss-qho}). This means that we only need to compute residual points but do not have to share any solution data. We utilize the distributed deep learning framework Horovod \cite{hvd}. 
The scalability analysis was done during the first 100 epochs on using 240 batches consisting of 35000 residual points each and 20 epochs for pretraining. The baseline network is a 8-layer MLP with 200 neurons per layer. Performance measurements were done by forking one benchmark process per compute node.

\begin{figure}[H]
\begin{center}
\includegraphics[width=0.7\textwidth]{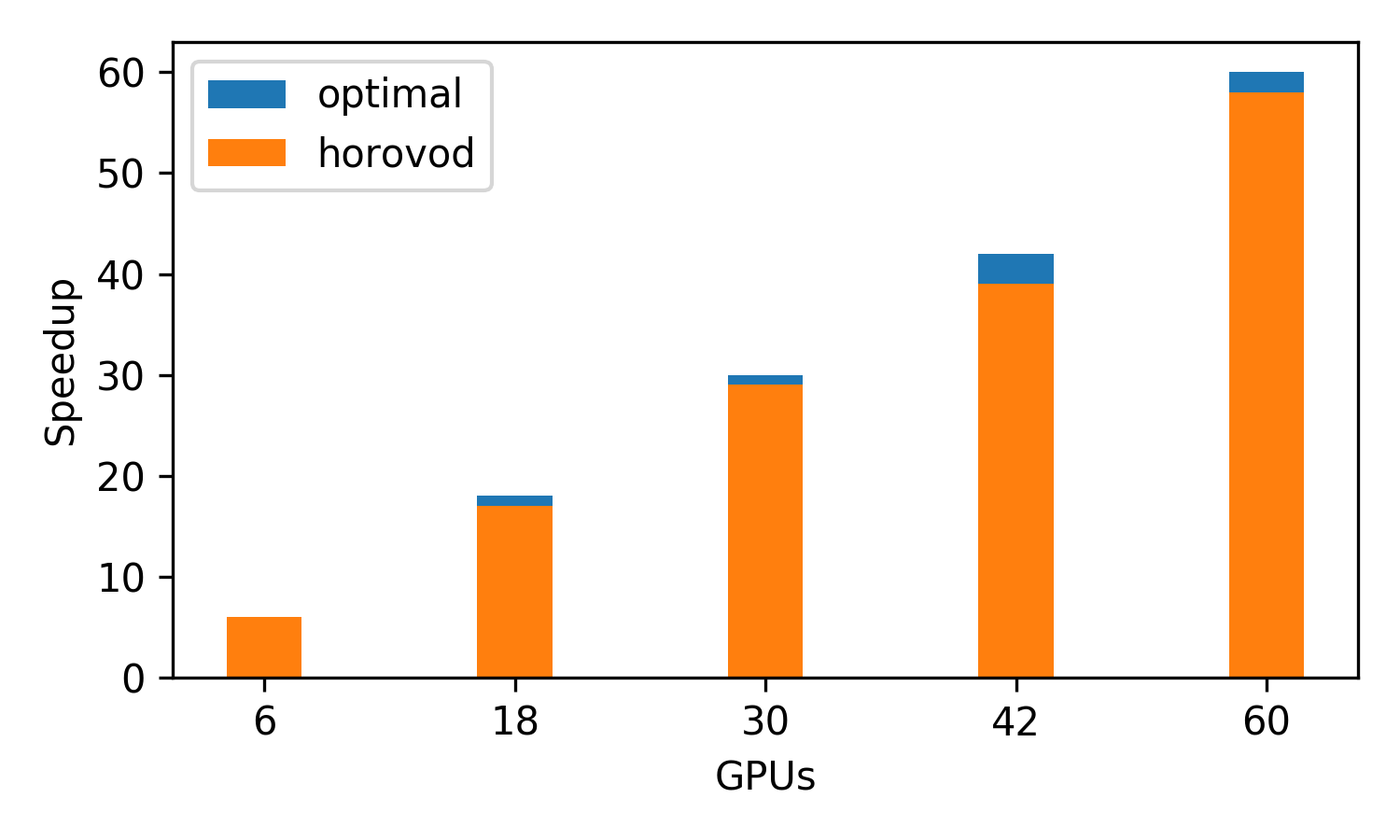}
\caption{Speedup comparison}
\label{fig:speedup}
\end{center}
\end{figure}

 Figure \ref{fig:speedup} compares the optimal with the actual speedup.
The speedup $S(k)$ for $k$-GPUs was computed by 
\begin{equation*}
S(k) = t_k / t_1 \ ,
\end{equation*}
provided the runtime for 100 epochs of a single GPU $t_1$ compared to the runtime of $k$ GPUs: $t_k$. We found almost linear speedup, though the the difference to the optimum is probably due to the latency of the communication between the GPUs and the distribution of residual points and gradient updates. The training achieved an average GPU utilization of $95 \% \pm 0.69 \% $ almost fully utilizing each GPU. Memory utilization stays relatively low at an average of $65 \% \pm 0.48 \%$  while most of the utilization relates to duplicates of the computational graph due to automatic differentiation. 


\begin{figure}[H]
\begin{center}
\includegraphics[width=0.7\textwidth]{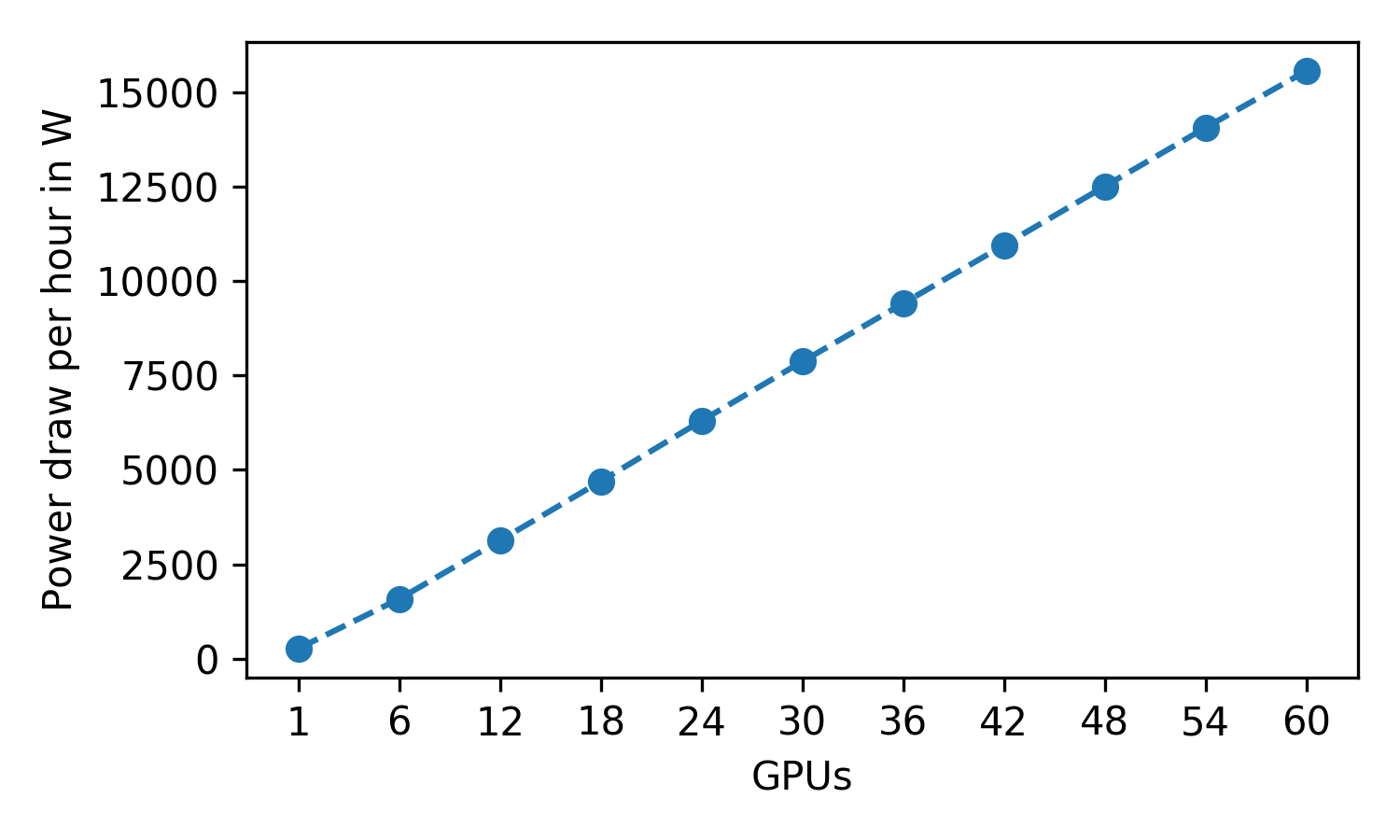}
\caption{Power draw comparison}
\label{fig:power_draw}
\end{center}
\end{figure}
We also quantified the power draw relating to the training in terms of the average hourly draw of all GPUs \ref{fig:power_draw}. Note that this rough measure omits the resting-state power draw of each compute node. We found an almost linear increase in power draw when increasing the number of GPUs. This correlates with the already mentioned very high GPU utilization as well as speedup. These findings imply that total energy for training our network for 100 epochs stays the same - no matter how many GPUs we use. 
Summarizing, Horovod has proven to be an excellent choice for the distributed training of physics-informed neural networks since training is compute bound. Note that the linear scalability has an upper bound caused by the time needed to perform the ring-allreduce and the splitting of the data.


\subsection{Discussion}
The experimental results of this paper agree with theoretical results on convergence of PINNs for parabolic and elliptic partial differential equations\cite{Shin2020} even for large two-dimensional problems such as the quantum harmonic oscillator. This benchmark dataset\footnote{The PyTorch implementations of the benchmarking dataset as well as the neural solvers for 1D and 2D Schrodinger equation and pretrained models are available online: \url{https://github.com/ComputationalRadiationPhysics/NeuralSolvers}} provides all means for a comprehensive assessment of approximation error as well as scalability due to the availability of an analytic solution while the smoothness of the solution can be altered by frequency $\omega$ of the QHO. The approximated solution of Physics-informed neural networks approached the quality of state-of-the-art spectral solvers for the QHO\cite{SpectralSolver}. The training time of PINN or GatedPINN is not competitive to the runtime of spectral solvers for \textit{one} 2D simulation. However, PINN enable warm-starting simulations by transfer learning techniques, integrating parameters (e.g. $\omega$ in our case) or Physics-informed solutions to inverse problems \cite{Chen2020} making that approach more flexible than traditional solvers. The former two approaches might tackle that challenge by learning complex relationships among parameters\cite{Michoski2019} or adapting a simulation to a new configuration at faster training time than learning it from scratch while the latter might pave the way for future  experimental usage. The GatedPINN architecture finally allows us to approach higher dimensional data when training physics-informed neural networks by training $k$ sub-PINN each representing a certain fraction of the computational domain at $1/k$ of the total PINN capacity. GatedPINN preserve the accuracy of PINN while the training time was reduced by 40\% (table 3). This effect will become even more evident for 3D or higher dimensional problems. Limiting the computational blowup of PINN and retaining linear speedup (see Fig. 7) are crucial steps towards the applications of physics-informed neural networks on e.g. three-dimensional or complex and coupled partial differential equations. 

\section{Conclusion}

Physics-informed neural networks denote a recent general purpose vehicle for machine learning assisted solving of partial differential equations. These neural solvers are solely trained on initial conditions while the time-dependent solution is recovered by solving an optimization problem. However, a major bottleneck of neural solvers is the high demand in capacity for representing the solution which relates to the size, dimension and complexity of the compute domain. In this work, we approach that issue by learning a domain decomposition and utilizing multiple tiny neural networks. GatedPINNs basically reduce the number of parameters per unit volume of our compute domain which reduces the training time while almost retaining the accuracy of the baseline neural solver. We find these results on a novel benchmark based on the 2D quantum harmonic oscillator. Additionally, GatedPINN estimate high-quality solutions of the physical system while the speedup is almost linear even for a large amount of GPUs.

\section*{Acknowledgement}
This work was partially funded by the Center of Advanced Systems Understanding (CASUS) which is financed by Germany’s Federal Ministry of Education and Research (BMBF) and by the Saxon Ministry for Science, Culture and Tourism (SMWK) with tax funds on the basis of the budget approved by the Saxon State Parliament. The authors gratefully acknowledge the GWK support for funding this project by providing computing time through the Center for Information Services and HPC (ZIH) at TU Dresden on the HPC-DA.
\bibliographystyle{unsrt}
\bibliography{library}

\begin{thebibliography}{10}

\bibitem{Vallecorsa2018}
S.~Vallecorsa.
\newblock {Generative models for fast simulation}.
\newblock {\em Journal of Physics: Conference Series}, 1085(2), 2018.

\bibitem{Raissi2018}
Maziar Raissi.
\newblock {Deep Hidden Physics Models: Deep Learning of Nonlinear Partial
  Differential Equations}.
\newblock {\em Journal of Machine Learning Research}, 19:1--24, 2018.

\bibitem{Raissi2017}
Maziar Raissi, Paris Perdikaris, and George~Em Karniadakis.
\newblock {Physics Informed Deep Learning ( Part I ): Data-driven Solutions of
  Nonlinear Partial Differential Equations}.
\newblock (Part I):1--22, 2017.

\bibitem{kim19deep}
Byungsoo Kim, Vinicius C.~Azevedo, Nils Thuerey, Theodore Kim, Markus Gross,
  and Barbara Solenthaler.
\newblock {Deep Fluids: A Generative Network for Parameterized Fluid
  Simulations}.
\newblock {\em Computer Graphics Forum (Proc. Eurographics)}, 38(2), 2019.

\bibitem{Tompson2017}
Jonathan Tompson, Kristofer Schlachter, Pablo Sprechmann, and Ken Perlin.
\newblock {Accelerating Eulerian Fluid Simulation With Convolutional Networks}.
\newblock In {\em Proc. 34th International Conference on Machine Learning},
  2017.

\bibitem{Shin2020}
Yeonjong Shin, Jerome Darbon, and George~Em Karniadakis.
\newblock {On the Convergence and generalization of Physics Informed Neural
  Networks}.
\newblock 02912:1--29, 2020.

\bibitem{PPINN}
Xuhui Meng, Zhen Li, Dongkun Zhang, and George~Em Karniadakis.
\newblock Ppinn: Parareal physics-informed neural network for time-dependent
  pdes, 2019.

\bibitem{MoE}
Noam Shazeer, Azalia Mirhoseini, Krzysztof Maziarz, Andy Davis, Quoc Le,
  Geoffrey Hinton, and Jeff Dean.
\newblock Outrageously large neural networks: The sparsely-gated
  mixture-of-experts layer, 2017.

\bibitem{conditionalComputing}
Emmanuel Bengio, Pierre-Luc Bacon, Joelle Pineau, and Doina Precup.
\newblock Conditional computation in neural networks for faster models, 2015.

\bibitem{hvd}
Alexander Sergeev and Mike~Del Balso.
\newblock {\em Horovod: fast and easy distributed deep learning in TensorFlow}.
\newblock 2018.

\bibitem{SpectralSolver}
Solution of the schrödinger equation by a spectral method.
\newblock {\em Journal of Computational Physics}, 47(3):412 -- 433, 1982.

\bibitem{Chen2020}
Yuyao Chen, Lu~Lu, George~Em Karniadakis, and Luca {Dal Negro}.
\newblock {Physics-informed neural networks for inverse problems in nano-optics
  and metamaterials}.
\newblock {\em Optics Express}, 28(8):11618, 2020.

\bibitem{Michoski2019}
Craig Michoski, Milos Milosavljevic, Todd Oliver, and David Hatch.
\newblock {Solving Irregular and Data-enriched Differential Equations using
  Deep Neural Networks}.
\newblock {\em CoRR}, 78712:1--22, 2019.

\end{thebibliography}

\end{document}